\documentclass[sigconf]{acmart}
\usepackage{amsmath}
\usepackage{courier}
\usepackage{bm}
\usepackage{algpseudocode}
\usepackage[noend,ruled,linesnumbered]{algorithm2e}
\usepackage{color}
\usepackage{enumitem}
\usepackage{url}
\usepackage{graphicx}
\usepackage{array}
\usepackage{multirow}
\usepackage{pifont}
\usepackage{xcolor,colortbl}
\usepackage{setspace}
\usepackage{subfigure}
\usepackage{makecell}
\usepackage{balance}

\newcommand{\ours}{\textsc{FuTex}\xspace}
\newcommand{\simm}{{\rm sim}}
\newcommand{\score}{{\rm score}}
\newcommand{\plm}{{\rm PLM}}
\newcommand{\bme}{\bm e}
\newcommand{\bmw}{\bm w}
\newcommand{\bmh}{\bm h}
\newcommand{\bmx}{\bm x}
\newcommand{\bmy}{\bm y}
\newcommand{\bmz}{\bm z}

\definecolor{myblue}{rgb}{0.2, 0.2, 0.9}
\definecolor{myorange}{rgb}{0.9, 0.5, 0.0}
\newcommand{\cmark}{\ding{51}}

\newcommand\blfootnote[1]{%
  \begingroup
  \renewcommand\thefootnote{}\footnote{#1}%
  \addtocounter{footnote}{-1}%
  \endgroup
}

\copyrightyear{2023}
\acmYear{2023}
\setcopyright{acmlicensed}\acmConference[KDD '23]{Proceedings of the 29th ACM SIGKDD Conference on Knowledge Discovery and Data Mining}{August 6--10, 2023}{Long Beach, CA, USA}
\acmBooktitle{Proceedings of the 29th ACM SIGKDD Conference on Knowledge Discovery and Data Mining (KDD '23), August 6--10, 2023, Long Beach, CA, USA}
\acmPrice{15.00}
\acmDOI{10.1145/3580305.3599544}
\acmISBN{979-8-4007-0103-0/23/08}

\begin{document}
\title{Weakly Supervised Multi-Label Classification of \\ Full-Text Scientific Papers}

\author{Yu Zhang}
\affiliation{
\institution{University of Illinois at Urbana-Champaign} 
\country{}
\institution{yuz9@illinois.edu}
}

\author{Bowen Jin}
\affiliation{
\institution{University of Illinois at Urbana-Champaign} 
\country{}
\institution{bowenj4@illinois.edu}
}

\author{Xiusi Chen}
\affiliation{
\institution{University of California, Los Angeles} 
\country{}
\institution{xchen@cs.ucla.edu}
}

\author{Yanzhen Shen}
\affiliation{
\institution{University of Illinois at Urbana-Champaign} 
\country{}
\institution{yanzhen4@illinois.edu}
}

\author{Yunyi Zhang}
\affiliation{
\institution{University of Illinois at Urbana-Champaign} 
\country{}
\institution{yzhan238@illinois.edu}
}

\author{Yu Meng}
\affiliation{
\institution{University of Illinois at Urbana-Champaign} 
\country{}
\institution{yumeng5@illinois.edu}
}

\author{Jiawei Han}
\affiliation{
\institution{University of Illinois at Urbana-Champaign} 
\country{}
\institution{hanj@illinois.edu}
}

\renewcommand{\shortauthors}{Yu Zhang et al.}

\begin{abstract}
Instead of relying on human-annotated training samples to build a classifier, weakly supervised scientific paper classification aims to classify papers only using category descriptions (e.g., category names, category-indicative keywords). Existing studies on weakly supervised paper classification are less concerned with two challenges: (1) Papers should be classified into not only coarse-grained research topics but also fine-grained themes, and potentially into multiple themes, given a large and fine-grained label space; and (2) full text should be utilized to complement the paper title and abstract for classification. Moreover, instead of viewing the entire paper as a long linear sequence, one should exploit the structural information such as citation links across papers and the hierarchy of sections and paragraphs in each paper. To tackle these challenges, in this study, we propose \ours, a framework that uses the cross-paper network structure and the in-paper hierarchy structure to classify full-text scientific papers under weak supervision. A network-aware contrastive fine-tuning module and a hierarchy-aware aggregation module are designed to leverage the two types of structural signals, respectively. Experiments on two benchmark datasets demonstrate that \ours significantly outperforms competitive baselines and is on par with fully supervised classifiers that use 1,000 to 60,000 ground-truth training samples.
\blfootnote{$^\dagger$Code and Datasets are available at {\color{myblue} \url{https://github.com/yuzhimanhua/FUTEX}}.}
\end{abstract}

\ccsdesc[500]{Information systems~Data mining}
\ccsdesc[500]{Computing methodologies~Classification and regression trees}

\keywords{multi-label text classification; weak supervision; scientific paper; full text}

\begin{spacing}{0.97}

\maketitle

\section{Introduction}
Weakly supervised text classification \cite{mekala2020contextualized,meng2020text,wang2021x} aims to classify text documents into a set of pre-defined categories without relying on any human-labeled training documents. Instead, the classifier seeks help from various formats of weak supervision such as category names \cite{meng2018weakly,meng2020text,wang2021x,shen2021taxoclass}, a few category-indicative keywords \cite{zhang2019higitclass,mekala2020contextualized,zhang2021weakly}, and category descriptions \cite{zhang2022metadata}. This setting significantly alleviates the burden of manual annotations, which is particularly helpful in some real applications such as scientific paper classification, where annotations need to be acquired from domain experts.

Although existing studies on weakly supervised text classification have applied their proposed methods to scientific paper datasets such as arXiv \cite{meng2019weakly,zhang2021hierarchical} and DBLP \cite{mekala2020meta,zhang2022motifclass}, they are less concerned with the following two challenges in practice.

\vspace{1mm}

\noindent \textbf{A Large and Fine-Grained Label Space.} One major goal of scientific paper classification is to help researchers track and analyze academic information and resources. To facilitate this goal, papers should be classified into not only coarse-grained research fields (e.g., ``\textsf{Machine Learning}'' and ``\textsf{Public Health}'') but also fine-grained themes (e.g., ``\textsf{Large Language Models}'' and ``\textsf{Deltacoronavirus}''). Note that in a large and fine-grained label space, most papers are naturally relevant to multiple themes. However, most existing studies under the weakly supervised setting focus on classifying papers at a coarse level with 5 to 50 categories and assume each document is relevant to only one category (or a single path from the root to a leaf if categories form a hierarchy). As far as we know, MICoL \cite{zhang2022metadata} is a pioneering work that considers weakly supervised multi-label classification with more than 10,000 categories. Nevertheless, its accuracy is hampered by using limited information such as paper titles and abstracts only, which will be discussed below.

\vspace{1mm}

\noindent \textbf{The Usage of Paper Full Texts.} A paper's title and abstract, although summarized to cover its major topics, cannot capture all fine-grained aspects. 
For example, technique-related labels may be introduced in more detail in the ``Method'' section; downstream tasks may be explained in the ``Experiments'' section. To find more labels relevant to a paper, it becomes necessary to leverage its full text. 
Intuitively, when we aim to utilize paper full texts, the first challenge is to deal with long text. 
Indeed, if we check the two datasets, MAG-CS and PubMed, used by MICoL \cite{zhang2022metadata}, the average full-paper length exceeds 4,000 words. In comparison, most pre-trained language models (PLMs) such as BERT \cite{devlin2019bert} and SciBERT \cite{beltagy2019scibert} can take an input sequence with at most 512 tokens. Therefore, those PLM-based weakly supervised text classifiers \cite{mekala2020contextualized,meng2020text,wang2021x,zhang2022metadata} cannot be directly applied to full-text scientific papers. 

\begin{figure}[!t]
\centering
\subfigure[MAG-CS]{
\includegraphics[width=0.23\textwidth]{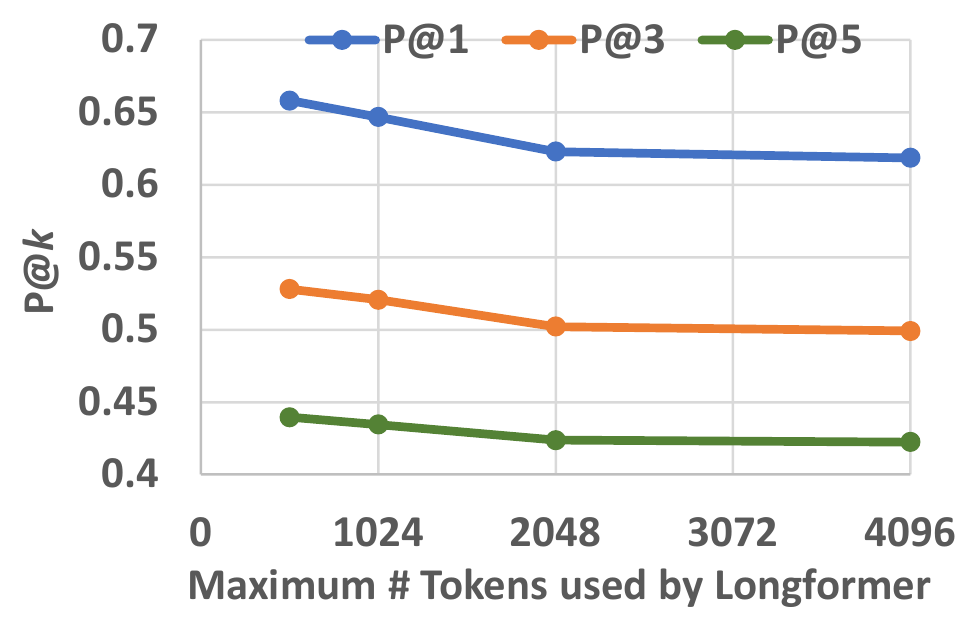}}
\hspace{-1mm}
\subfigure[PubMed]{
\includegraphics[width=0.23\textwidth]{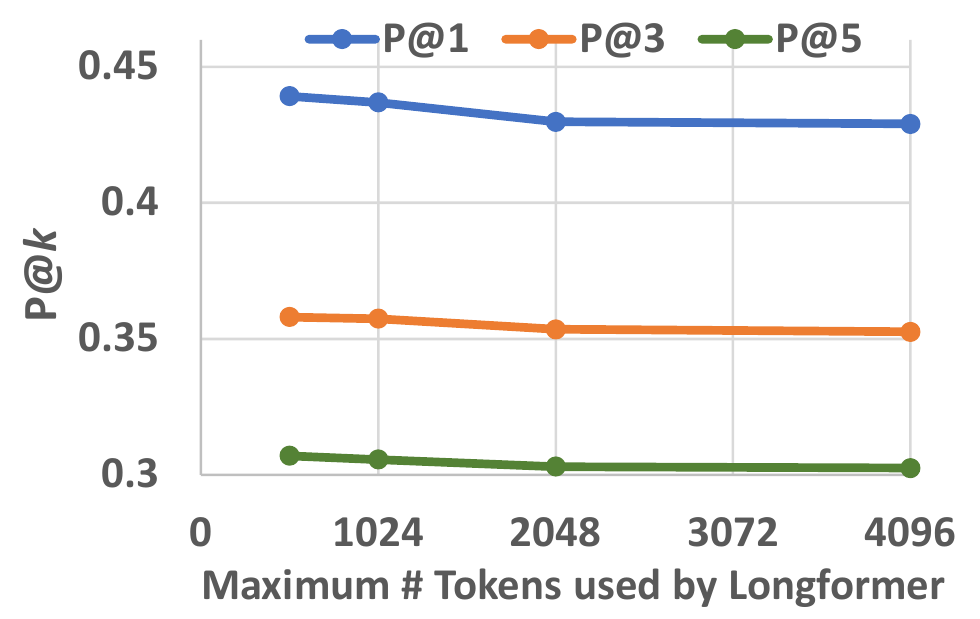}}
\vspace{-1em}
\caption{Weakly supervised classification performance of Longformer \cite{beltagy2019scibert} on two datasets of scientific papers, MAG-CS and PubMed, used in \cite{zhang2021match}. When Longformer allows a larger input sequence length (i.e., it can take more tokens from paper full texts), the classification precision drops.} 
\vspace{-0.5em}
\label{fig:longformer}
\end{figure}

\begin{figure}[!t]
\centering
\includegraphics[width=0.98\linewidth]{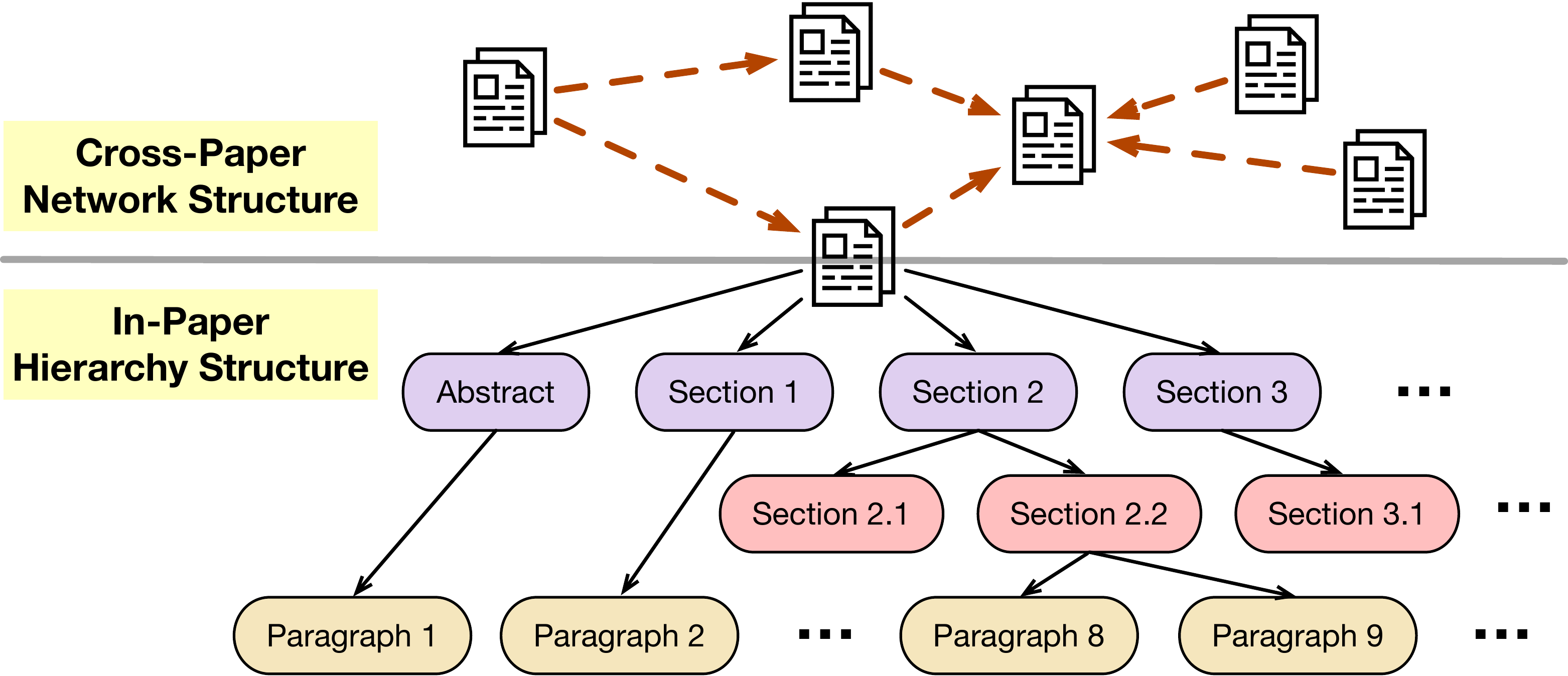}
\vspace{-0.5em}
\caption{The cross-paper network structure and in-paper hierarchy structure associated with scientific papers.}
\vspace{-0.5em}
\label{fig:intro}
\end{figure}

More importantly, we would like to argue that classifying full-text papers is beyond the problem of dealing with long text. As shown in Figure \ref{fig:longformer}, we adopt Longformer \cite{beltagy2020longformer}, which can take at most 4,096 tokens, to classify papers in MAG-CS and PubMed under the weakly supervised setting. 
(For more experiment details such as how Longformer is used, please refer to Section \ref{sec:setup}.) 
We control the maximum input sequence length of Longformer from 512 to 4,096. Surprisingly, the more tokens Longformer takes (i.e., the more information from the full text that Longformer considers), the lower the classification precision is. 
This observation has two implications: First, full texts are noisy and should not be treated the same as abstracts. Abstracts should still play a leading role in classification, while full texts provide auxiliary information \cite{jimeno2013mesh,dai2020fullmesh}. Second, to better exploit full text, we need to consider the structures in scientific papers. The major design that enables Longformer to take a longer input sequence is to sparsify the fully connected attention in Transformer \cite{vaswani2017attention}, where each input token only interacts with its neighbor tokens and the first several tokens in the linear sequence. 
However, the rich structural information prevalently available inside the full text is not fully captured by Longformer.
Figure \ref{fig:intro} shows two types of such structural information: the \textit{cross-paper network structure} and the \textit{in-paper hierarchy structure}. The hierarchy structure organizes sections, subsections, and paragraphs into a tree, where the parent-child relation indicates that a finer text unit is included in a coarser one. Such a structure implies which paragraphs should be jointly considered when aggregating paragraph semantics to the entire paper. Moreover, by parsing the bibliographic entries in paper full texts, one can obtain each paper's references and construct a citation network. This cross-paper network structure indicates the semantic proximity between two papers. To summarize, these two types of structures provide additional semantic signals that are not reflected in a linear text sequence.

\vspace{1mm}

\noindent \textbf{Contributions.} Being aware of the two aforementioned challenges, in this paper, we study weakly supervised multi-label text classification of full-text scientific papers. We propose \ours, the design of which is centered around how to use the cross-paper network structure and the in-paper hierarchy structure to classify scientific papers in a large and fine-grained label space. \ours has three major modules: (1) \textit{Network-aware contrastive fine-tuning} aims to leverage the cross-paper network structure to fine-tune a pre-trained language model (PLM) so that it can probe fine-grained label semantics and distinguish among similar categories. (2) \textit{Hierarchy-aware aggregation} aims to exploit the in-paper hierarchy structure to obtain the entire paper representation by aggregating from its paragraphs. With this aggregation process, a PLM does not need to deal with the full-text sequence at once. (3) \textit{Self-training} aims to take the initial prediction (i.e., top-ranked categories according to the first two modules) as pseudo labels to train a full-text paper classifier. Then the prediction of the trained classifier can complement the initial prediction to improve the final classification results.

We conduct experiments on two datasets (both with $>$10,000 categories) commonly used in previous studies \cite{zhang2021match,ye2021beyond,zhang2022metadata}. Results show that \ours outperforms competitive baselines including scientific PLMs \cite{beltagy2019scibert,liu2022oag,cohan2020specter}, weakly supervised text classifiers \cite{yin2019benchmarking,zhang2022metadata}, and structure-enhanced PLMs \cite{yasunaga2022linkbert,yang2021graphformers}. Notably, on the MAG-CS dataset, our \ours model, without any ground-truth training data, is on par with a supervised classifier \cite{prabhu2018parabel} trained on 60,000 labeled papers. To summarize, this work makes the following contributions.
\begin{itemize}[leftmargin=*]
\item We study the problem of weakly supervised multi-label classification of full-text scientific papers. Different from most previous studies on weakly supervised text classification, it considers a large, fine-grained label space and paper full texts.
\item We propose the \ours framework which utilizes the cross-paper network structure and the in-paper hierarchy structure associated with scientific papers to improve classification performance.
\item We conduct experiments on two datasets and demonstrate the effectiveness of \ours in comparison with competitive baselines, including those using abstracts only and those using full text.
\end{itemize}

\section{Preliminaries}
\subsection{Cross-Paper and In-Paper Structures}
\noindent \textbf{Cross-Paper Network Structure.} Given a collection of scientific papers, we can obtain the references of each paper by parsing the bibliographic entries in its full text. If we view papers as nodes and references as directed edges, a network can be constructed. Formally, we have the definition below.

\begin{definition}{(Cross-Paper Network Structure)}
A scientific paper corpus $\mathcal{D}$ has a cross-paper network structure $\mathcal{G}=(\mathcal{V}, \mathcal{E})$, where each node $d \in \mathcal{V}$ is a paper, and $(d_i, d_j) \in \mathcal{E}$ if and only if $d_i$ cites $d_j$ (i.e., $d_j$ is a bibliographic entry in $d_i$).
\label{def:network}
\end{definition}

To describe the relationship between two papers in $\mathcal{G}$, we adopt the notation of meta-paths \cite{sun2011co,sun2011pathsim}. To be specific, if $d_i$ cites $d_j$, we can say the two papers are connected via the meta-path ``Paper $\rightarrow$ Paper'' (or its abbreviation $P\rightarrow P$). Similarly, if two papers $d_i$ and $d_j$ share a common reference $d_k$ (i.e., $(d_i, d_k) \in \mathcal{E}$ and $(d_j, d_k) \in \mathcal{E}$), we can say $d_i$ and $d_j$ are connected via the meta-path $P\rightarrow P\leftarrow P$. Intuitively, if two papers are connected via a certain meta-path, their relevant topics are more likely to overlap. Following \cite{zhang2022metadata}, given a meta-path $\mathcal{M}$, we use $d_i \rightarrow^{\mathcal{M}} d_j$ to denote that $d_i$ is connected to $d_j$ via $\mathcal{M}$, and the meta-path-based neighborhood $\mathcal{N}_{\mathcal{M}}(d_i)$ is defined as $\{d_j|d_i \rightarrow^{\mathcal{M}} d_j\ {\rm AND}\ d_j \neq d_i\}$.

\vspace{1mm}

\noindent \textbf{In-Paper Hierarchy Structure.} One unique challenge we are facing in full-text paper classification is that each paper is beyond a plain text sequence (i.e., title+abstract) and contains its internal hierarchical structure of paragraphs. As shown in Figure \ref{fig:intro}, nodes representing paragraphs, subsections, sections, and the entire paper form a tree, in which the parent-child relation implies a finer text unit is entailed by a coarser one. Formally, we have the definition below. 

\begin{definition}{(In-Paper Hierarchy Structure)}
A full-text paper $d$ contains a hierarchical tree structure $\mathcal{T}_d$. The root of $\mathcal{T}_d$ represents the entire paper; the leaves of $\mathcal{T}_d$ are $d$'s paragraphs $\mathcal{P}_d=\{p_{d1},...,p_{dM}\}$. The tree can be characterized by a mapping $\textsf{Child}(\cdot)$, where $\textsf{Child}(x)$ is the set of text units that are one level finer than $x$ and contained in $x$.
\label{def:hierarchy}
\end{definition}

Putting the two structures together, we would like to classify the nodes in a network $\mathcal{G}$. Meanwhile, each node $d$ contains its own subcomponents that form a hierarchy $\mathcal{T}_d$. Given a paper, we need to jointly consider its subcomponents and its proximity with neighbors to infer its categories.

\subsection{Problem Definition}
\begin{figure}[!t]
\centering
\includegraphics[width=\linewidth]{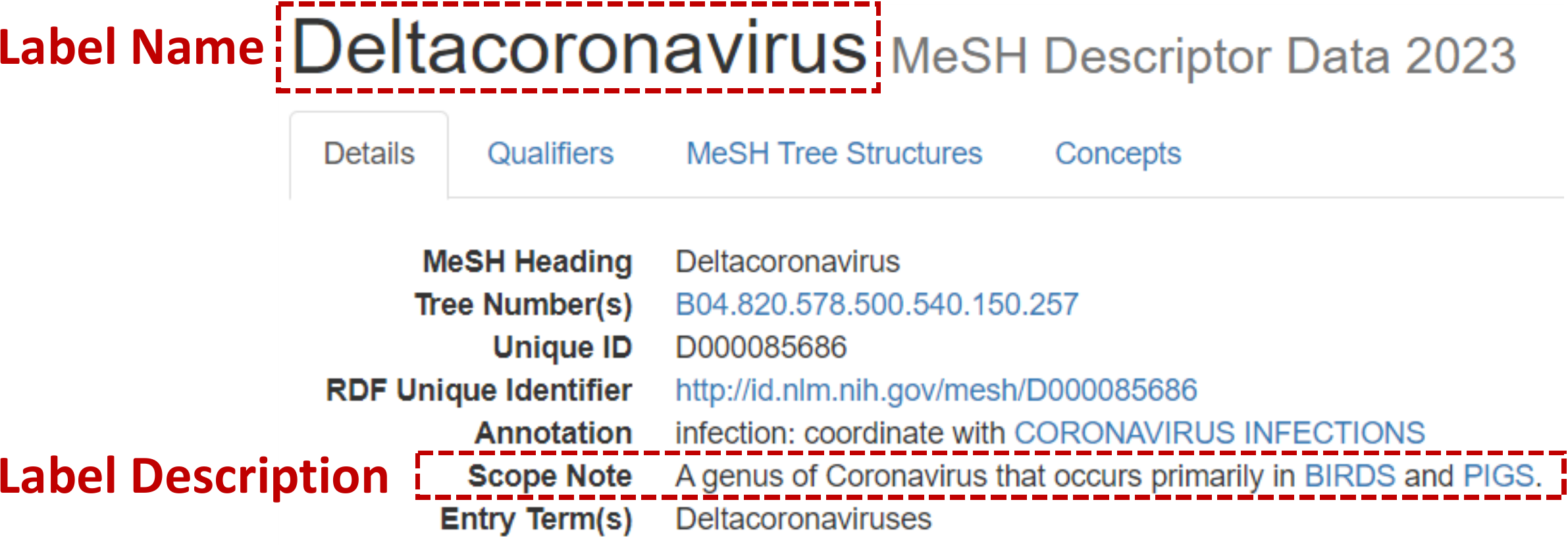}
\vspace{-1.2em}
\caption{Name and description of the label ``\textsf{Deltacorona- virus}'' from PubMed (\url{https://meshb.nlm.nih.gov/record/ui?ui=D000085686}).}
\vspace{-0.7em}
\label{fig:label}
\end{figure}

In this paper, we study weakly supervised multi-label text classification. By ``weakly supervised'', we imply that we do not have any annotated training samples for any label, and the only available supervision to characterize a label is its name and several descriptive sentences \cite{chai2020description,zhang2022metadata}. Figure \ref{fig:label} shows the name and description of the label ``\textsf{Deltacoronavirus}'' as an example. This setting is more challenging than zero-shot multi-label text classification \cite{gupta2021generalized,nam2015predicting,rios2018few,xiong2022extreme,zhang2019integrating} which assumes annotated documents are given for some \textit{seen} classes and the trained classifier should be generalized to predict \textit{unseen} classes. Under the weakly supervised setting, all classes are unseen. This setting is also called ``\textit{wild} zero-shot'' in some previous studies \cite{yin2019benchmarking,zhang2022metadata}.

By ``multi-label'', we mean that each paper can be relevant to more than one label. This is a natural assumption when the label space is fine-grained and multi-faceted. For example, a COVID-19 paper can be labeled as ``\textsf{Infections}'', ``\textsf{Lung Diseases}'', ``\textsf{Coronavirus}'', and ``\textsf{Public Health}'' at the same time. This assumption makes our task more challenging than weakly supervised single-label classification \cite{mekala2020contextualized,meng2020text,wang2021x,zhang2021weakly,park2022lime}.

To summarize, our task can be defined as follows.
\begin{definition}{(Problem Definition)}
Given (1) a collection of scientific papers $\mathcal{D}$ with a cross-paper network structure $\mathcal{G}$, where each paper $d$ has its full text and hierarchy structure $\mathcal{T}_d$, and (2) a label space $\mathcal{L}$ where each label $l$ has its name and description, our task is to predict the relevant labels $\mathcal{L}_d \subseteq \mathcal{L}$ for each $d \in \mathcal{D}$.
\label{def:problem}
\end{definition}

\section{Model}
 \begin{figure*}[!t]
\centering
\includegraphics[width=\textwidth]{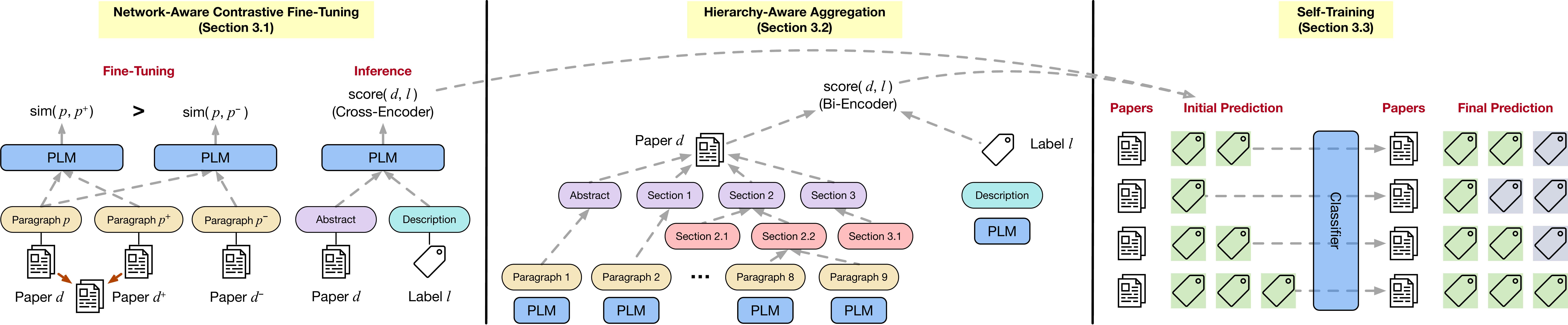}
\vspace{-1.2em}
\caption{Overview of the \ours framework.}
\vspace{-0.5em}
\label{fig:framework}
\end{figure*}

One straightforward solution to our task is to pick a pre-trained language model (e.g., SciBERT \cite{beltagy2019scibert}), use it to encode each paper's content and each label's name/description to get their embeddings, and then perform the nearest neighbor search in the embedding space. However, such an approach suffers from two drawbacks: First, unfine-tuned PLMs may not be powerful enough to detect the subtle semantic differences between two papers or two label descriptions, but fine-grained text classification, to a great extent, requires the classifier to distinguish among labels that are close to each other. Second, the entire paper is long (e.g., with $\sim$4,000 words on average in the Semantic Scholar Open Research Corpus (S2ORC) \cite{lo2020s2orc}), which exceeds the maximum sequence length (e.g., 512 tokens) that a PLM can handle in most cases.

To overcome the aforementioned two drawbacks, we propose to exploit the cross-paper network structure and the in-paper hierarchy structure, which will be introduced in Sections \ref{sec:network} and \ref{sec:hierarchy}, respectively. Then, in Section \ref{sec:self}, we present a self-training strategy, that is, how we use initial predictions as pseudo labels to train a classifier that complements the predictions. The overview of our proposed \ours framework is shown in Figure \ref{fig:framework}.

\subsection{Network-Aware Contrastive Fine-Tuning}
\label{sec:network}

The first module in \ours aims to utilize the cross-paper network structure to improve the PLM's ability to distinguish among fine-grained labels. We follow the intuition of LinkBERT \cite{yasunaga2022linkbert} that if two papers are connected via certain citation-based relationships (e.g., $d_i \rightarrow^{\mathcal{M}} d_j$), then a paragraph $p_s \in d_i$ and a paragraph $p_t \in d_j$ are more likely to share fine-grained topics than two randomly picked paragraphs. In LinkBERT \cite{yasunaga2022linkbert}, Yasunaga et al. propose to concatenate the two ``linked'' paragraphs together (i.e., $\texttt{[CLS]} p_s \texttt{[SEP]} p_t \texttt{[SEP]}$) to perform masked token prediction and document relation prediction for language model pre-training \cite{devlin2019bert}. However, in this paper, we are not aiming at training a general-purpose PLM. Instead, our model only needs to judge whether two text units are relevant to similar topics or not. Being able to do this, during inference, the model can take a paragraph and a label description as input to predict whether the paragraph is relevant to the label. To achieve this goal, following \cite{zhang2022metadata}, we adopt a contrastive fine-tuning objective to replace the language model pre-training objectives in LinkBERT.

To be specific, the PLM should be fine-tuned to distinguish between ``linked'' and ``unlinked'' paragraph pairs. As shown in Figure \ref{fig:framework} (left), given three papers $d$, $d^+$, and $d^-$, where $d^+\in \mathcal{N}_{\mathcal{M}}(d)$ and $d^-\notin \mathcal{N}_{\mathcal{M}}(d)$. We randomly sample three paragraphs $p$, $p^+$, and $p^-$ from $d$, $d^+$, and $d^-$, respectively. The PLM aims to predict the similarity between $p$ and $p^+$ as well as that between $p$ and $p^-$.
\begin{equation}
\begin{split}
\bme_{\rm pos} &= \plm(\texttt{[CLS]} p \texttt{[SEP]} p^+ \texttt{[SEP]}), \\
\bme_{\rm neg} &= \plm(\texttt{[CLS]} p \texttt{[SEP]} p^- \texttt{[SEP]}).
\end{split}
\end{equation}
Here, $\bme_{\rm pos}$ and $\bme_{\rm neg}$ are the output representations of the \texttt{[CLS]} token after PLM encoding.

A linear layer is then trained to predict the scores given the output representations.
\begin{equation}
\simm(p, p^+) = \bmw^\top \bme_{\rm pos}, \ \ \ \simm(p, p^-) = \bmw^\top \bme_{\rm neg},
\end{equation}
where $\bmw$ is a learnable vector.
The PLM is fine-tuned to make $\simm(p, p^+)$ larger than $\simm(p, p^-)$, in which case we can adopt the contrastive loss \cite{chen2020simple}.
\begin{equation}
    \mathcal{J} = \mathop{\mathbb{E}}_{\substack{p \in d \\ p^+ \in d^+ \in \mathcal{N}_{\mathcal{M}}(d) \\ p^- \in d^- \notin \mathcal{N}_{\mathcal{M}}(d)}} \bigg[-\log \frac{\exp(\simm(p, p^+))}{\exp(\simm(p, p^+))+\exp(\simm(p, p^-))}\bigg].
\label{eqn:contrastive}
\end{equation}

After contrastive fine-tuning, the PLM is used to predict the score between a paragraph and a label. For example, given a paper $d \in \mathcal{D}$ and a label $l \in \mathcal{L}$, we use $a_d$ to denote the title+abstract of $d$, and we use $t_l$ to denote the name+description of $l$. Then, the score can be calculated as
\begin{equation}
    \simm(a_d, t_l) = \bmw^\top \plm(\texttt{[CLS]} a_d \texttt{[SEP]} t_l \texttt{[SEP]}).
\label{eqn:cross}
\end{equation}

However, the strategy that concatenates each paragraph and each label and feeds them into one PLM (i.e., the Cross-Encoder architecture \cite{nogueira2019multi}) makes the inference computationally expensive because the representation of each text unit cannot be pre-computed. For example, suppose there are $D$ papers (each of which has $P$ paragraphs) and $L$ labels, then we need to call the PLM $O(DPL)$ times during inference. Such a cost will prohibit us from applying the model to a large corpus and a large label space.  For example, if $D=L=10^4$ and $P=30$, then $DPL=3\times 10^9$. To alleviate the cost, we adopt the following two strategies.

\vspace{1mm}

\noindent \textbf{Adding a retrieval stage.} Following \cite{zhang2022metadata}, given a paper $d$, we first adopt exact name matching to retrieve a small set of candidate labels $\mathcal{C}(d)$ from the entire label space $\mathcal{L}$. To be specific, if a label's name appears in the paper's content, it will be added as a candidate. The fine-tuned PLM is then applied as a reranker to score labels from the retrieved candidate pool.

\vspace{1mm}

\noindent \textbf{Using Bi-Encoder for non-abstract paragraphs.} As mentioned in the Introduction, the abstract and other paragraphs should not be treated equally in text classification. The abstract is highly summarized to cover the major topics of the paper, while a paragraph in the paper body may capture only one aspect and provides auxiliary topic signals. To use the more powerful tool in the most important case, we only adopt the Cross-Encoder architecture when inferring the labels of paper abstracts (i.e., Eq. (\ref{eqn:cross})). For other paragraphs, we adopt the Bi-Encoder architecture \cite{karpukhin2020dense}. To be specific, we encode each paragraph and each label separately.
\begin{equation}
    \bmh_p = \plm(\texttt{[CLS]} p \texttt{[SEP]}), \ \ \ 
    \bmh_l = \plm(\texttt{[CLS]} t_l \texttt{[SEP]}).
\label{eqn:bi}
\end{equation}
Here, $\bmh_p$ and $\bmh_l$ are the output representations of the \texttt{[CLS]} token after PLM encoding.
Then, the similarity between $p$ and $l$ can be computed as $\cos(\bmh_p, \bmh_l)$. One drawback of this strategy is that the paragraph and the label text cannot serve as each other's context during PLM encoding. However, the efficiency is significantly improved because we can pre-compute $\bmh_p$ and $\bmh_l$ for all paragraphs and labels. In fact, if we combine the two proposed strategies, the inference complexity will be reduced to $O(D\lambda+DP+L)$, where $\lambda$ is the average number of candidate labels picked for each paper in the retrieval stage. For example, if we assume $D=L=10^4$ and $P=\lambda=30$, then $D\lambda+DP+L=6.1\times 10^5$, which is orders of magnitude smaller than $DPL=3\times 10^9$.

\subsection{Hierarchy-Aware Aggregation}
\label{sec:hierarchy}

Although the similarity between each paragraph $p$ and each label $l$ can be computed efficiently now, we have not figured out how to calculate the score between an entire paper $d$ and a label $l$. Intuitively, simply averaging all paragraph embeddings may not work well because important signals from the paper abstract and conclusion will then be buried under the great amount of content from other sections. Previously, FullMeSH \cite{dai2020fullmesh} proposed to check 5 sections -- abstract, introduction, method, result, and summary -- in each full paper so as to probe relevant topics from different aspects. Inspired by their idea, we utilize the in-paper hierarchy structure $\mathcal{T}_d$ to perform embedding aggregation from paragraphs to sections, and then to the entire paper.

Given a non-leaf text unit $x \in \mathcal{T}_d$ (e.g., subsections, sections, or the entire paper), we obtain the embedding of $x$ by aggregating the embeddings from $x$'s children. Formally,
\begin{equation}
   \bmh_x = \frac{1}{|\textsf{Child}(x)|} \sum_{y\in \textsf{Child}(x)} \bmh_y.
\label{eqn:aggr}
\end{equation}
After the bottom-up aggregation, the score between the entire paper 

\noindent $d$ and the label $l$ can be computed as
\begin{equation}
   \score_B(d, l) = \cos(\bmh_d, \bmh_l).
\label{eqn:scoreb}
\end{equation}
Here, the subscript ``$B$'' means the score is based on the Bi-Encoder architecture. Recall that we can calculate the score between $d$ and $l$ using Eq. (\ref{eqn:cross}) based on the Cross-Encoder architecture:
\begin{equation}
   \score_X(d, l) = \bmw^\top \plm(\texttt{[CLS]} a_d \texttt{[SEP]} t_l \texttt{[SEP]}),
\label{eqn:scorex}
\end{equation}
where ``$X$'' stands for Cross-Encoder.

Now we adopt an ensemble ranking step to jointly consider the two scores. Given a document $d$, we first rank all candidate labels from $\mathcal{C}(d)$ in descending order according to $\score_B(d, l)$ and $\score_X(d, l)$, respectively. In this way, each candidate label $l$ will have two rank positions $r_B(l|d)$ and $r_X(l|d)$. Then, we calculate the mean reciprocal rank (MRR) of $l$.
\begin{equation}
    {\rm MRR}(l|d) = \frac{1}{r_B(l|d)} + \frac{1}{r_X(l|d)}.
\label{eqn:mrr}
\end{equation}
Finally, candidate labels are sorted according to ${\rm MRR}(l|d)$ as the reranking result.

\subsection{Self-Training}
\label{sec:self}
The retrieval stage proposed in Section \ref{sec:network} significantly improves the efficiency of \ours. However, it may filter out labels that do not explicitly appear in a paper but are implicitly relevant to the paper in the latent semantic space. To mitigate this issue, we present a self-training strategy to utilize paper full texts and confident predictions (based on MRR) to train a text classifier $f_{\rm class}(\cdot)$. The classifier is then used to predict the probability that a paper is relevant to a label (not necessarily selected in the retrieval stage), and those top-ranked labels will complement our initial MRR-based predictions.

Since the major goal of this paper is not to invent a new fully supervised text classifier, we propose to use an off-the-shelf model.
To leverage paper full texts and meanwhile bypass the sequence length problem, we choose a bag-of-words multi-label classifier -- Parabel \cite{prabhu2018parabel}, which, according to \cite{you2019attentionxml}, has competitive performance even compared with deep learning classifiers on benchmark datasets.
Parabel represents each training document $d$ as a $|\mathcal{W}_\mathcal{D}|$-dimensional feature vector $\bmx_d$, where $\mathcal{W}_\mathcal{D}$ is the vocabulary of $\mathcal{D}$. Given a word $w \in \mathcal{W}_\mathcal{D}$, its corresponding entry in $\bmx_d$ is the following tf--idf score:
\begin{equation}
    x_{d,w} = {\rm tf}(w, d) \cdot {\rm idf}(w, \mathcal{D}),
\label{eqn:selfx}
\end{equation}
where ${\rm tf}(w, d)$ is the term frequency of $w$ in $d$, and ${\rm idf}(w, \mathcal{D}) =$ $\log\frac{|\mathcal{D}|}{|\{d'\in \mathcal{D}|w\in d'\}|}$ is the inverse document frequency of $w$.

For each paper $d$, according to MRR calculated in Eq. (\ref{eqn:mrr}), we use the top-$N$ predicted labels as pseudo labels to train the Parabel classifier. (If $d$ has less than $N$ candidate labels selected in the retrieval stage, we use all of them.) Formally, the pseudo labels of $d$ is represented as an $|\mathcal{L}|$-dimensional vector $\bmy_d$, where
\begin{equation}
y_{d,l} = 
\begin{cases}
1, & r_{\rm MRR}(l|d) \leq N, \\
0,  & {\rm otherwise}.
\end{cases}
\label{eqn:selfy}
\end{equation}
Here, $r_{\rm MRR}(l|d)$ is the rank position of $l$ according to ${\rm MRR}(l|d)$.

Given $\bmx_d$ and $\bmy_d$ of each paper, Parabel learns a tree-based discriminative classifier $\Pr(\hat{\bmy}_d|\bmx_d)$. (For more technical details, one can refer to \cite{prabhu2018parabel}.) Based on the trained classifier, we relabel each paper $d$ to keep those initial confident predictions while incorporating new highly-ranked labels. To be specific, the top-$N$ predictions according to MRR remain at top-$N$ in our final predictions; the other labels will be sorted according to Parabel's output $\Pr(\hat{y}_{dl}=1|\bmx_d)$ and ranked after top-$N$. We use the following example to explain this process.

\begin{example}{(Final Prediction after Self-Training)}
Suppose there are 5 labels $\mathcal{L}=\{A, B, C, D, E\}$ and $N=2$.
Given a paper $d$, assume 3 labels $A$, $B$, and $C$ are selected in the retrieval stage and their MRR scores are $2.00$, $1.00$, and $0.67$, respectively. Then, $A$ and $B$ will be the two pseudo labels of $d$ used for training Parabel. The trained Parabel is used to classify $d$ again and get the following scores:

\begin{table}[H]
\centering
\vspace{-0.5em}
\scalebox{0.86}{
\begin{tabular}{c|ccccc}
\hline
$l$ & $A$ & $B$ & $C$ & $D$ & $E$ \\ \hline
$\Pr(\hat{y}_{dl}=1|\bmx_d)$ & 0.80 & 0.85 & 0.30 & 0.60 & 0.90 \\
\hline
\end{tabular}
}
\vspace{-0.5em}
\end{table}

\noindent In our final prediction, $A$ and $B$ will still be the top 2, and the other labels will be ranked according to Parabel's prediction. Therefore, the final rank will be $(A, B, E, D, C)$. In practice, we find this strategy achieves better classification performance than reranking all labels purely based on $\Pr(\hat{y}_{dl}=1|\bmx_d)$, the result of which is $(E, B, A, D, C)$.

\vspace{-0.5em}
\label{ex:final}
\end{example}

The entire procedure of \ours is summarized in Appendix \ref{sec:pseudocode}.

\section{Experiments}
\subsection{Setup}
\label{sec:setup}
\noindent \textbf{Datasets.} We use two datasets, \textbf{MAG-CS} \cite{wang2020microsoft} and \textbf{PubMed} \cite{lu2011pubmed}, that are widely adopted in previous studies on scientific paper classification \cite{zhang2021match,ye2021beyond,zhang2022metadata}. Originally, MAG-CS contains $\sim$705K papers published at 105 top computer science venues, where each paper is labeled with its related fields-of-study \cite{shen2018web}; PubMed consists of $\sim$899K papers published in 150 top medicine journals, where each paper is labeled with its related MeSH terms \cite{coletti2001medical}. However, since the two original datasets do not have paper full texts, we try to extract full texts from S2ORC \cite{lo2020s2orc}. S2ORC has segmented each paper into paragraphs, marked the section that each paragraph belongs to, and parsed the bibliographic entries. We remove the paragraphs with less than 10 words from each paper. Note that not all papers in these two datasets can be found in S2ORC, and we finally obtain 96,718 full-text MAG-CS papers and 251,573 full-text PubMed papers. Because \ours does not require any annotated training data, all these papers are used for testing. More statistics of these two datasets can be found in Table \ref{tab:dataset}.

\begin{table}[!t]
\centering
\caption{Dataset Statistics}
\vspace{-0.5em}
\scalebox{0.86}{
\begin{tabular}{c|ccccc}
\hline
Dataset & \#Papers & \#Labels 
& \begin{tabular}[c]{@{}c@{}}\#Words/ \\ Paper\end{tabular}
& \begin{tabular}[c]{@{}c@{}}\#Paragraphs/ \\ Paper\end{tabular} 
& \begin{tabular}[c]{@{}c@{}}\#Labels/ \\ Paper\end{tabular} \\ \hline
MAG-CS \cite{wang2020microsoft}  & 96,718    & 10,909    & 4071.69       & 45.91         & 5.84           \\
PubMed \cite{lu2011pubmed}  & 251,573   & 16,070    & 4901.42       & 33.38         & 8.69         \\ \hline 
\end{tabular}
}
\vspace{-1em}
\label{tab:dataset}
\end{table}

\vspace{1mm}

\noindent \textbf{Compared Methods.} We compare our \ours model with the following baselines including scientific PLMs, structure-enhanced PLMs, and zero-shot multi-label text classification methods.

\begin{table*}[!t]
\centering
\caption{P@$k$ and NDCG@$k$ scores of compared methods on MAG-CS and PubMed. Bold: the highest score. *: \ours is significantly better than this method with p-value $< 0.05$. **: \ours is significantly better than this method with p-value $< 0.01$.}
\scalebox{0.9}{
\begin{tabular}{l|c|ccccc|ccccc}
\hline
& \multirow{2}{*}{\textbf{Method}} & \multicolumn{5}{c|}{\textbf{MAG-CS \cite{wang2020microsoft}}}  & \multicolumn{5}{c}{\textbf{PubMed \cite{lu2011pubmed}}}  \\ \cline{3-12}
&                            & \textbf{P@1}             & \textbf{P@3}             & \textbf{P@5}             & \textbf{NDCG@3}          & \textbf{NDCG@5}          & \textbf{P@1}             & \textbf{P@3}             & \textbf{P@5}             & \textbf{NDCG@3}          & \textbf{NDCG@5}          \\ \hline
{\multirow{6}{*}{\rotatebox[origin=c]{90}{Abstract}}} & SciBERT \cite{beltagy2019scibert}        & 0.6825**       & 0.5525**       & 0.4542**       & 0.5990**          & 0.5555**          & 0.4503**       & 0.3712**       & 0.3193**       & 0.3916**          & 0.3595**          \\
& OAG-BERT \cite{liu2022oag}       & 0.5960**       & 0.4860**       & 0.4168**       & 0.5258**          & 0.5018**          & 0.4142**       & 0.3340**       & 0.2902**       & 0.3539**          & 0.3265**          \\
& LinkBERT \cite{yasunaga2022linkbert}       & 0.6515**       & 0.5288**       & 0.4402**       & 0.5729**          & 0.5363**          & 0.3689**       & 0.3331**       & 0.3014**       & 0.3434**          & 0.3270**          \\
& SPECTER \cite{cohan2020specter}        & 0.7599**       & 0.5958**       & 0.4765**       & 0.6510**          & 0.5923**          & 0.5419**       & 0.4124**       & 0.3388**       & 0.4441**          & 0.3946**          \\
& 0SHOT-TC \cite{yin2019benchmarking,liu2019roberta}       & 0.6489**       & 0.5003**       & 0.4164**       & 0.5490**          & 0.5128**          & 0.5504**       & 0.4280**       & 0.3532**       & 0.4583**          & 0.4085**          \\
& MICoL \cite{zhang2022metadata}          & 0.7658       & 0.6005**       & 0.4797**       & 0.6562**          & 0.5965**          & 0.5775*       & 0.4329**       & 0.3528**       & 0.4680**          & 0.4133**          \\ \hline
{\multirow{7}{*}{\rotatebox[origin=c]{90}{Full Text}}} & SciBERT \cite{beltagy2019scibert}        & 0.6881**       & 0.5573**       & 0.4574**       & 0.6041**          & 0.5594**          & 0.4885**       & 0.4176**       & 0.3517**       & 0.4375**          & 0.3965**          \\
& OAG-BERT \cite{liu2022oag}       & 0.5775**       & 0.4724**       & 0.4097**       & 0.5108**          & 0.4913**          & 0.4245**       & 0.3867**       & 0.3296**       & 0.4002**          & 0.3660**          \\
& LinkBERT \cite{yasunaga2022linkbert}       & 0.6654**       & 0.5365**       & 0.4447**       & 0.5820**          & 0.5429**       & 0.4241**       & 0.3919**       & 0.3380**       & 0.4041**       & 0.3727**       \\
& SPECTER \cite{cohan2020specter}        & 0.7582**       & 0.5965**       & 0.4773**       & 0.6512**          & 0.5927**          & 0.5400**       & 0.4335**       & 0.3575**       & 0.4610**          & 0.4112**          \\
& Longformer \cite{beltagy2020longformer}     & 0.6582**       & 0.5280**       & 0.4396**       & 0.5740**          & 0.5366**          & 0.4392**       & 0.3580**       & 0.3070**       & 0.3786**          & 0.3466**          \\
& PLM+GAT \cite{yang2021graphformers,velickovic2018graph}    & 0.7306** & 0.5760** & 0.4663** & 0.6285** & 0.5765** & 0.5140** & 0.4000** & 0.3374** & 0.4279** & 0.3870**   \\
& GraphFormers \cite{yang2021graphformers}   & 0.7261** & 0.5720** & 0.4637** & 0.6242** & 0.5730** & 0.5213** & 0.4046** & 0.3380** & 0.4334** & 0.3895**   \\ \hline
& \ours           & \textbf{0.7692}       & \textbf{0.6089}       & \textbf{0.4914}       & \textbf{0.6648}          & \textbf{0.6099}          & \textbf{0.5900}       & \textbf{0.4533}       & \textbf{0.3798}       & \textbf{0.4869}          & \textbf{0.4388}  \\ \hline       
\end{tabular}
}
\label{tab:performance}
\end{table*}

\begin{itemize}[leftmargin=*]
\item \textbf{SciBERT \cite{beltagy2019scibert}}\footnote{\url{https://huggingface.co/allenai/scibert_scivocab_uncased}} is a scientific PLM trained on 1.14M scientific papers from Semantic Scholar \cite{ammar2018construction} using masked language modeling and next sentence prediction tasks.
\item \textbf{OAG-BERT \cite{liu2022oag}}\footnote{\url{https://github.com/THUDM/OAG-BERT}} is a scientific PLM trained on 120M scientific papers from the Open Academic Graph \cite{zhang2019oag}. It proposes heterogeneous entity type embedding, span-aware entity masking, and entity-aware two-dimensional position embedding to leverage paper metadata information (e.g., venues and authors).
\item \textbf{LinkBERT \cite{yasunaga2022linkbert}}\footnote{\url{https://huggingface.co/michiyasunaga/LinkBERT-base}} is a structure-enhanced PLM. It uses linked Wikipedia paragraphs as context to perform masked language modeling and document relation prediction. On the PubMed dataset, we use BioLinkBERT\footnote{\url{https://huggingface.co/michiyasunaga/BioLinkBERT-base}}, which is pre-trained on PubMed papers and performs better than LinkBERT.
\item \textbf{SPECTER \cite{cohan2020specter}}\footnote{\url{https://huggingface.co/allenai/specter}} is a structure-enhanced scientific PLM. It continues pre-training SciBERT using a citation prediction objective with 684K pairs of linked papers.
\end{itemize}
The four baselines above can be used to classify either abstracts or full-text papers. When applying them to abstracts, we directly encode each abstract and each label description to calculate the cosine similarity between the two embeddings. When applying them to full text, we follow the hierarchy-aware aggregation process in Section \ref{sec:hierarchy} and ensemble the results of using abstract only and using full text.

\begin{itemize}[leftmargin=*]
\item \textbf{0SHOT-TC \cite{yin2019benchmarking,liu2019roberta}} is a zero-shot text classification method. It is a natural language inference (NLI) model that predicts to what extent a paper (as the premise) entails the sentence ``\textit{this document is about \{label\_name\}}.'' (as the hypothesis). Following \cite{shen2021taxoclass}, we use \textbf{RoBERTa-large-mnli} \cite{liu2019roberta}\footnote{\url{https://huggingface.co/roberta-large-mnli}} as the NLI model.
\item \textbf{MICoL \cite{zhang2022metadata}}\footnote{\url{https://github.com/yuzhimanhua/MICoL}} is a zero-shot text classification method. It proposes a metadata-induced contrastive learning technique to fine-tune a PLM. MICoL has various configurations with the model architecture and the used meta-path. According to the experimental results in \cite{zhang2022metadata}, we choose the best-performing configuration: (Cross-Encoder, $P\rightarrow P\leftarrow P$).
\end{itemize}
The two methods above adopt the Cross-Encoder architecture. As mentioned in Section \ref{sec:network}, it will be too computationally expensive to apply them to all paragraphs. Therefore, we keep their original usage on paper abstracts only.

\begin{itemize}[leftmargin=*]
\item \textbf{Longformer \cite{beltagy2020longformer}}\footnote{\url{https://huggingface.co/allenai/longformer-base-4096}} is a PLM dealing with long documents. It sparsifies the fully connected attention and can take 4,096 tokens at most. We adopt it to encode full-text papers by setting the maximum number of tokens as 512, 1,024, 2,048, and 4,096. As shown in the Introduction, the 512-token version performs the best, so we use it for performance comparison.
\item \textbf{PLM+GAT \cite{yang2021graphformers,velickovic2018graph}} is a PLM stacked with a Graph Attention Network (GAT) layer. Each paragraph $p \in d$ is first encoded by a PLM. Then we use GAT to obtain paragraph embeddings by aggregating PLM representations of its neighbor paragraphs $p' \in d^+ \in \mathcal{N}_{\mathcal{M}}(d)$. A link prediction objective (i.e., judging whether two paragraphs are connected via meta-path $\mathcal{M}$) is then adopted to train the PLM and GAT in an end-to-end manner.
\item \textbf{GraphFormers \cite{yang2021graphformers}}\footnote{\url{https://github.com/microsoft/GraphFormers}} is a GNN-nested PLM architecture, in which GNN layers and Transformer layers are alternately stacked. Similar to PLM+GAT, a link prediction objective is adopted to train the model so that the PLM can be enhanced by the cross-paper network structure.
\end{itemize}
The three methods above are naturally suitable for classifying paper full texts. 

According to our experiments, SPECTER performs better than SciBERT, OAG-BERT, and LinkBERT. Therefore, for MICoL, PLM+ GAT, GraphFormers, and \ours, we all use SPECTER as the base PLM. Also, the meta-path $\mathcal{M}$ is set as $P\rightarrow P\leftarrow P$ for all these models for a fair comparison. 

\vspace{1mm}

\noindent \textbf{Implementation and Hyperparameters.}
Each label from the PubMed dataset may have multiple label names, including one canonical name and 0, 1, or several synonyms (i.e., ``entry terms''). Following \cite{zhang2022metadata}, we include $l$ as a candidate label of $d$ if \textit{any} of its names appears in $d$'s content. On both MAG-CS and PubMed, during the retrieval stage, we use each paper's title and abstract, instead of the full text, for label name matching because this yields better classification performance.
During network-aware contrastive fine-tuning, when feeding the two paragraphs into a Cross-Encoder, the maximum length of each paragraph is 256 tokens. The training batch size is 8. We use the AdamW optimizer \cite{loshchilov2019decoupled}, warm up the learning rate for the first 100 steps and then linearly decay it. The learning rate is 5e-5, the weight decay is 0.01, and $\epsilon=$ 1e-8. We sample 45,000 and 150,000 tuples of $(p, p^+, p^-)$ from MAG-CS and PubMed, respectively, for PLM fine-tuning.
During self-training, we set $N=5$ to get pseudo labels of each paper. For the Parabel classifier\footnote{http://manikvarma.org/code/Parabel/download.html}, all parameters are set by default. Specifically, the number of trees is 3; the maximum number of labels in a leaf node is 100; the beam search width in prediction is 10. We remove words appearing in less than 5 papers when training Parabel.

\vspace{1mm}

\noindent \textbf{Evaluation Metrics.} We use two commonly adopted metrics in multi-label text classification: P$@k$ and NDCG$@k$, where $k=1$, 3, and 5. Given a paper $d$, we use $\bmz_d \in \{0,1\}^{|\mathcal{L}|}$ to denote its ground-truth labels and ${\rm rank}(i)$ to denote the $i$-th ranked label in the final prediction. P$@k$ and NDCG$@k$ are then defined as:
\begin{equation}
\small
\begin{split}
    {\rm P@}k &= \frac{1}{k} \sum_{i = 1}^k z_{d, {\rm rank}(i)}. \\
    {\rm DCG@}k = \sum_{i=1}^k \frac{z_{d, {\rm rank}(i)}}{\log(i+1)}, \ \ \ &{\rm NDCG@}k = \frac{{\rm DCG@}k}{\sum_{i=1}^{\min(k, ||\bmz_d||_0)}\frac{1}{\log(i+1)}}.
\end{split}
\end{equation}

\subsection{Performance Comparison}
Table \ref{tab:performance} shows P$@k$ and NDCG$@k$ scores of compared methods on MAG-CS and PubMed. For models with randomness (i.e., MICoL, PLM+GAT, GraphFormers, and \ours), we run each of them 5 times with the average performance reported. Other models (i.e., SciBERT, OAG-BERT, LinkBERT, SPECTER, 0SHOT-TC, and Longformer) are deterministic according to our usage. To show statistical significance, we conduct a two-tailed t-test to compare \ours with each baseline if the baseline has randomness, and we conduct a two-tailed Z-test to compare \ours with each deterministic baseline. The significance level is also marked in Table \ref{tab:performance}.

From Table \ref{tab:performance}, we find that: (1) \ours consistently and significantly outperforms all baselines. On both datasets, MICoL is a competitive baseline, possibly because it also uses citation links across papers. However, since MICoL only considers text content from paper titles and abstracts, it is not as powerful as \ours. (2) There are four baselines (i.e., SciBERT, OAG-BERT, LinkBERT, and SPECTER) that can be used in both abstract-only and full-text settings. In most cases, a full-text variant can outperform its abstract-only counterpart. This observation validates our claim that considering the full text is beneficial to paper classification. Besides, unlike Longformer (as shown in Figure \ref{fig:longformer}), our proposed hierarchy-aware aggregation strategy, which is also used by the full-text variants of the four baselines, can effectively use paper full texts. Furthermore, in a few cases, a full-text variant underperforms its abstract-only counterpart in terms of P@1 but achieves higher P@3 and P@5. This finding implies that signals extracted by hierarchy-aware aggregation from paper full texts can better help lower-ranked predictions.

\vspace{1mm}

\begin{figure}[!t]
\centering
\subfigure[MAG-CS]{
\includegraphics[width=0.23\textwidth]{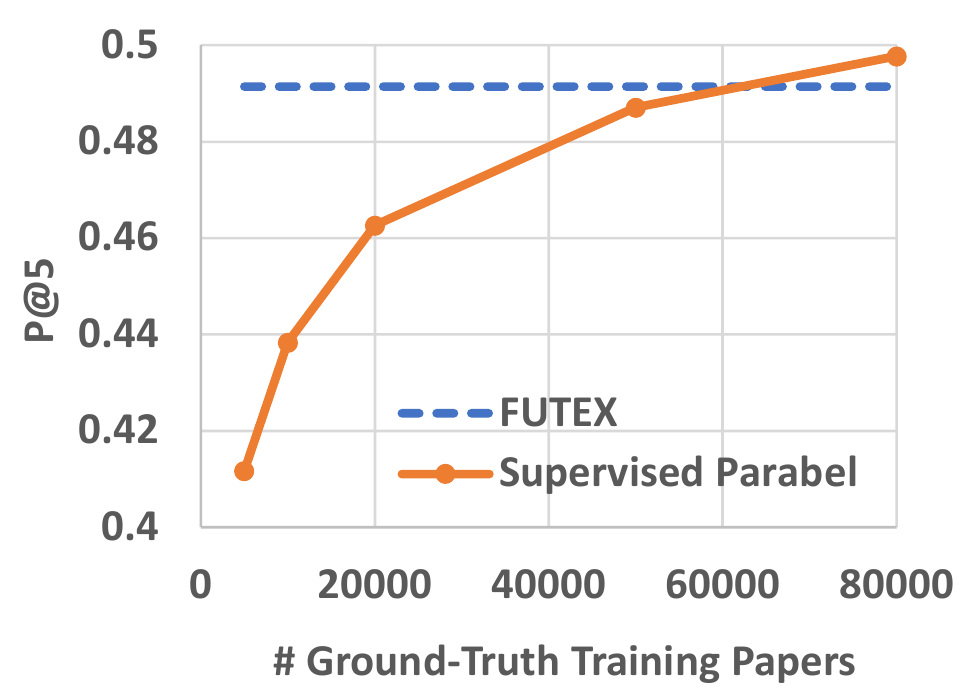}}
\hspace{-1mm}
\subfigure[PubMed]{
\includegraphics[width=0.23\textwidth]{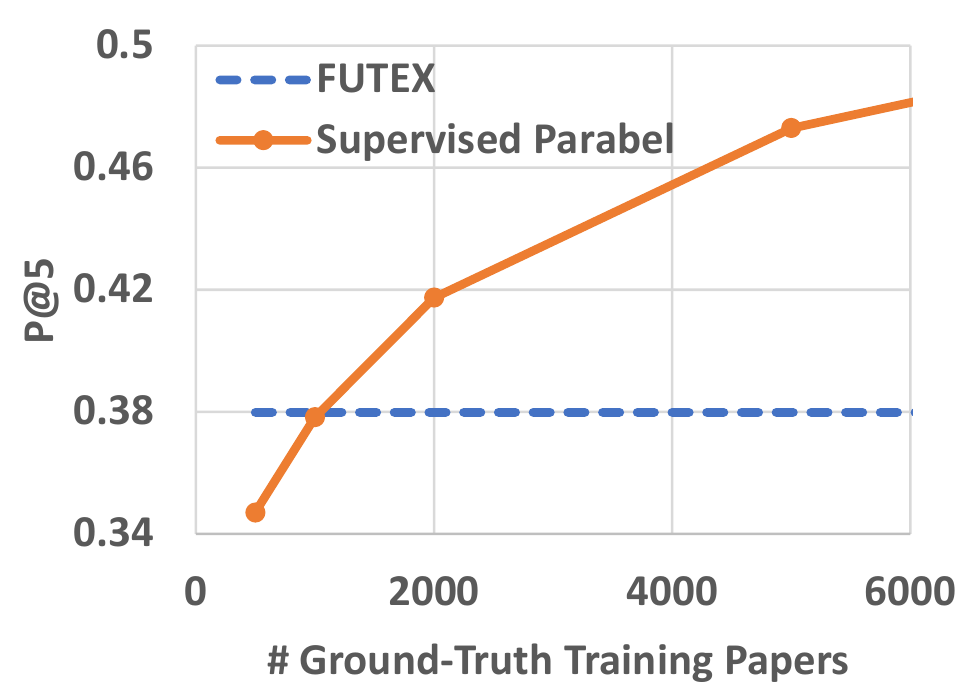}}
\vspace{-1em}
\caption{The P@5 score of supervised Parabel with different numbers of ground-truth training papers. Our \ours model, without any ground-truth training samples, is on par with Parabel that uses 60,000 and 1,000 training samples on MAG-CS and PubMed, respectively.} 
\vspace{-1em}
\label{fig:super}
\end{figure}

\noindent \textbf{Comparison with a Supervised Model.} We further compare \ours with a fully supervised multi-label text classifier. In accordance with our choice in Section \ref{sec:self}, we report the performance of Parabel \cite{prabhu2018parabel} with ground-truth training data. To be specific, for each dataset, we select 10,000 papers as testing samples and pick different numbers of training samples from the remaining papers. Figure \ref{fig:super} shows the P$@5$ score of Parabel with different numbers of ground-truth training samples in comparison with \ours. We can observe that our \ours model, without relying on any annotated training data, is on par Parabel that uses 60,000 and 1,000 ground-truth training samples on MAG-CS and PubMed, respectively.

\subsection{Ablation Study}
\begin{table*}[!t]
\centering
\caption{P@$k$ and NDCG@$k$ scores of the full \ours model and three ablation versions on MAG-CS and PubMed. Bold, *, and **: the same meaning as in Table \ref{tab:performance}.}
\scalebox{0.9}{
\begin{tabular}{c|ccccc|ccccc}
\hline
\multirow{2}{*}{\textbf{Method}} & \multicolumn{5}{c|}{\textbf{MAG-CS \cite{wang2020microsoft}}}  & \multicolumn{5}{c}{\textbf{PubMed \cite{lu2011pubmed}}}  \\ \cline{2-11}
& \textbf{P@1}             & \textbf{P@3}             & \textbf{P@5}             & \textbf{NDCG@3}          & \textbf{NDCG@5}          & \textbf{P@1}             & \textbf{P@3}             & \textbf{P@5}             & \textbf{NDCG@3}          & \textbf{NDCG@5}          \\ \hline
\ours-NoNetwork   & 0.7601** & 0.6013** & 0.4859** & 0.6567** & 0.6031** & 0.5422** & 0.4410** & 0.3722** & 0.4676** & 0.4244** \\
\ours-NoHierarchy & 0.7655** & 0.6048** & 0.4879** & 0.6607** & 0.6060** & 0.5802** & 0.4421** & 0.3681** & 0.4760** & 0.4272** \\
\ours-NoSelfTrain & 0.7673** & 0.6040** & 0.4827** & 0.6591** & 0.5993** & 0.5877** & 0.4456** & 0.3648** & 0.4801** & 0.4254** \\ \hline
\ours             & \textbf{0.7692} & \textbf{0.6089} & \textbf{0.4914} & \textbf{0.6648} & \textbf{0.6099} & \textbf{0.5900} & \textbf{0.4533} & \textbf{0.3798} & \textbf{0.4869} & \textbf{0.4388}   \\ \hline     
\end{tabular}
}
\label{tab:ablation}
\end{table*}

There are three major modules in \ours: network-aware contrastive fine-tuning, hierarchy-aware aggregation, and self-training. Now we conduct an ablation study to check the contribution of each module. To facilitate this, we create three ablation versions of \ours:
\begin{itemize}[leftmargin=*]
\item \textbf{\ours-NoNetwork} does not have the network-aware contrastive fine-tuning module. It directly uses SPECTER in the Bi-Encoder architecture, ensembles the results of using abstract only and using full text, and then performs self-training.
\item \textbf{\ours-NoHierarchy} does not have the hierarchy-aware aggregation module. After contrastive fine-tuning, it applies the Cross-Encoder to paper titles/abstracts only and then performs self-training.
\item \textbf{\ours-NoSelfTrain} does not have the self-training module. It uses the MRR-based ranking list obtained in Section \ref{sec:hierarchy} as the final prediction.
\end{itemize}

Table \ref{tab:ablation} demonstrates the performance of the full \ours model and the three ablation versions. We can observe that: (1) The full \ours model consistently and significantly outperforms the three ablation versions, indicating that all three major modules have a positive contribution to the classification performance. (2) Among the three ablation versions, \ours-NoNetwork performs the worst in terms of P@1. This finding indicates that the cross-paper network structure is more beneficial to top-ranked predictions. By contrast, \ours-NoSelfTrain has the lowest P@5 score on both datasets, which means that the self-training module contributes the most to lower-ranked predictions. This observation validates our claim that self-training can find more labels that are semantically relevant to each paper so as to complement the initial top-ranked categories.

\subsection{Case Study}
\begin{table*}[!t]
\centering
\footnotesize
\caption{Case study on MAG-CS and PubMed. {\color{myblue} Blue}: Labels indicated by the paper title or abstract. {\color{myorange} Orange}: Labels indicated by the paper's full text but not mentioned in the title or abstract.}
\vspace{-0.5em}
\begin{tabular}{p{0.48\textwidth} | p{0.48\textwidth}}
\hline
\multicolumn{1}{c|}{\textbf{MAG-CS \cite{wang2020microsoft}}} & \multicolumn{1}{c}{\textbf{PubMed \cite{lu2011pubmed}}} \\ \hline
\textbf{Title}: compact modeling technique for outdoor navigation & \textbf{Title}: serum \textbf{\color{myblue} calprotectin}: a novel diagnostic and prognostic marker in \textbf{\color{myblue} inflammatory bowel diseases} \\ \hline
\textbf{Abstract}: abstract-in this paper, a new methodology to build compact local maps in real time for outdoor \textbf{\color{myblue} robot} navigation is presented. the environment information is obtained from a 3-d \textbf{\color{myblue} scanner} laser. the navigation model, which is called traversable region model, is based on a \textbf{\color{myblue} voronoi diagram} technique ... & \textbf{Abstract}: there is an unmet need for novel blood-based biomarkers that offer timely and accurate diagnostic and prognostic testing in \textbf{\color{myblue} inflammatory bowel diseases} (ibd). we aimed to investigate the diagnostic and prognostic utility of serum \textbf{\color{myblue} calprotectin} (sc) in ibd ... \\ \hline
\textbf{Full Text}: this new challenge has prompted a change in robotics navigation philosophy, where \textbf{\color{myorange} path planning} and modeling were always obtained a priori ... in general, in \textbf{\color{myorange} mobile robot navigation}, the occupancy-based approach is one of the most commonly used methods ... \textbf{\color{myorange} path planning} in large and outdoor environments is a complex task because there are a lot of parameters that define the traversability, for example, as follows ... when the 3-d model is defined (as the one in figs. 5 and 6 , where the terrain considered ta is represented in blue and the nta is represented in red), the free space can be extracted to build a trm for the robot navigation and \textbf{\color{myorange} path planning} ... & \textbf{Full Text}: inflammatory bowel diseases (ibd), including \textbf{\color{myorange} crohn's disease} (cd) and \textbf{\color{myorange} ulcerative colitis} (uc), are chronic, debilitating inflammatory disorders of the gastrointestinal tract affecting adults and children ... a recent meta-analysis of 13 studies and 1,041 patients found that fc had a pooled \textbf{\color{myorange} sensitivity and specificity} of 0.93 ... to determine the accuracy of blood parameter measurements as a prognostic test capable of diagnosing ibd, receiver operating characteristic (roc) analyses were performed by plotting \textbf{\color{myorange} sensitivity} against \textbf{\color{myorange} specificity} ... \\ \hline \hline
\textbf{Ground-Truth Labels}: \textsf{robot}, \textsf{voronoi diagram}, \textsf{motion planning}, \textsf{computer vision}, \textsf{mobile robot navigation}, \textsf{mobile robot}, \textsf{scanner}, \textsf{computational geometry} & \textbf{Ground-Truth Labels}: \textsf{leukocyte l1 antigen complex}, \textsf{inflammatory bowel diseases}, \textsf{multivariate analysis}, \textsf{colitis ulcerative}, \textsf{prognosis}, \textsf{kaplan meier estimate}, \textsf{sensitivity and specificity}, \textsf{humans}, \textsf{crohn disease}, \textsf{logistic models}, \textsf{proportional hazards models}, \textsf{odds ratio}, \textsf{area under curve} \\ \hline
\textbf{MICoL Prediction}: \textsf{\color{myblue} voronoi diagram} (\cmark), \textsf{\color{myblue} robot} (\cmark), \textsf{\color{myblue} scanner} (\cmark) & \textbf{MICoL Prediction}: \textsf{\color{myblue} inflammatory bowel diseases} (\cmark), \textsf{\color{myblue} leukocyte l1 antigen complex} (\cmark) \\ \hline
\textbf{\ours Prediction}: \textsf{\color{myblue} voronoi diagram} (\cmark), \textsf{\color{myblue} robot} (\cmark), \textsf{\color{myblue} scanner} (\cmark), \textsf{\color{myorange} motion planning} (\cmark), \textsf{\color{myorange} mobile robot navigation} (\cmark) & \textbf{\ours Prediction}: \textsf{\color{myblue} inflammatory bowel diseases} (\cmark), \textsf{\color{myblue} leukocyte l1 antigen complex} (\cmark), \textsf{\color{myorange} crohn disease} (\cmark), \textsf{\color{myorange} colitis ulcerative} (\cmark), \textsf{\color{myorange} sensitivity and specificity} (\cmark) \\ \hline
\end{tabular}
\label{tab:case}
\end{table*}

We now perform a case study to qualitatively demonstrate the effect of considering full text in paper classification. Table \ref{tab:case} shows two cases, one of which is from MAG-CS and the other from PubMed. For both papers, we show their text information including the title, abstract, and excerpts from full text. We also show their ground-truth labels, labels predicted by \ours, and labels predicted by MICoL (which is the most competitive baseline using titles and abstracts only). We mark a label as {\color{myblue} blue} if it (or a semantically similar term) appears in the paper title or abstract; we mark a label as {\color{myorange} orange} if it does not appear in the title/abstract but is mentioned in full text.

In the MAG-CS case, three labels ``\textsf{voronoi diagram}'', ``\textsf{robot}'', and ``\textsf{scanner}'' explicitly appear in the paper abstract, and they are correctly predicted by both MICoL and \ours. However, MICoL misses labels such as ``\textsf{motion planning}'' and ``\textsf{mobile robot navigation}'', which are not mentioned in the title/abstract. In fact, the term ``outdoor robot navigation'' in the abstract may imply the paper's relevance to ``\textsf{mobile robot navigation}'', but MICoL does not build the connection between them. After the paper's full text is exploited, ``\textsf{mobile robot navigation}'' completely appears in the content, and the term ``path planning'', which is semantically close to the label ``\textsf{motion planning}'', is repeatedly mentioned. As a result, both labels are accurately captured by \ours. 

In the PubMed case, MICoL successfully predicts the ground-truth labels ``\textsf{inflammatory bowel diseases}'' and ``\textsf{leukocyte l1 antigen complex}'' (whose synonym is ``calprotectin''\footnote{According to \url{https://meshb-prev.nlm.nih.gov/record/ui?ui=D039841}, ``Calprotectin'' is an entry term of ``Leukocyte L1 Antigen Complex''.}) indicated by the title/abstract. However, MICoL fails to predict labels such as ``\textsf{crohn disease}'', ``\textsf{colitis ulcerative}'', and ``\textsf{sensitivity and specificity}''. As shown in Table \ref{tab:case}, hints to these labels can be found in the full text. Note that these labels are indeed relevant to the paper rather than just being mentioned: Crohn's disease and ulcerative colitis are two types of inflammatory bowel diseases studied in the paper, and the paper extensively discusses the sensitivity and specificity of predicting these diseases. By leveraging the paper's full text, \ours accurately picks these labels.

\subsection{Efficiency}
\begin{figure}[!t]
\centering
\subfigure[Training time]{
\includegraphics[width=0.23\textwidth]{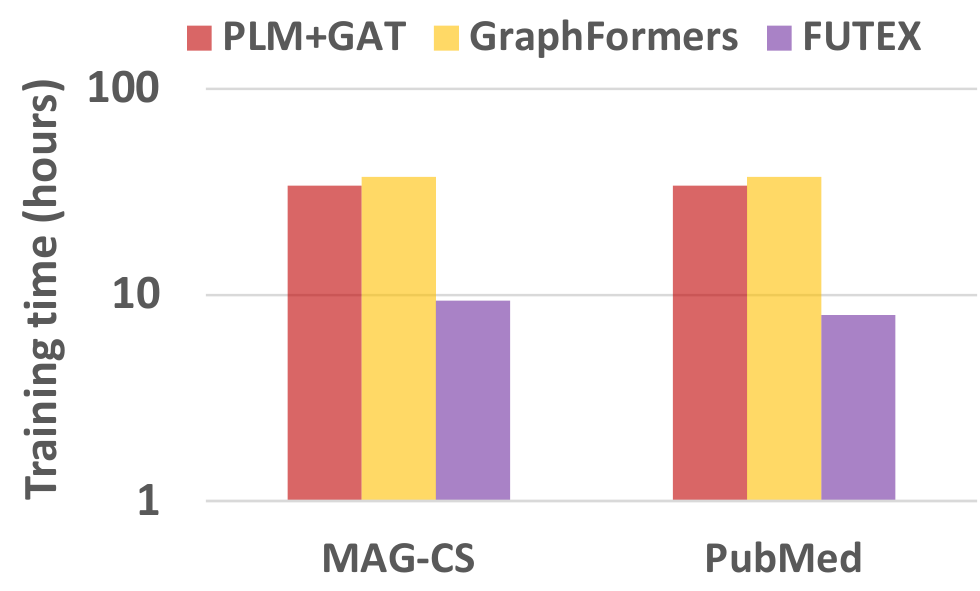}}
\hspace{-1mm}
\subfigure[Inference time]{
\includegraphics[width=0.23\textwidth]{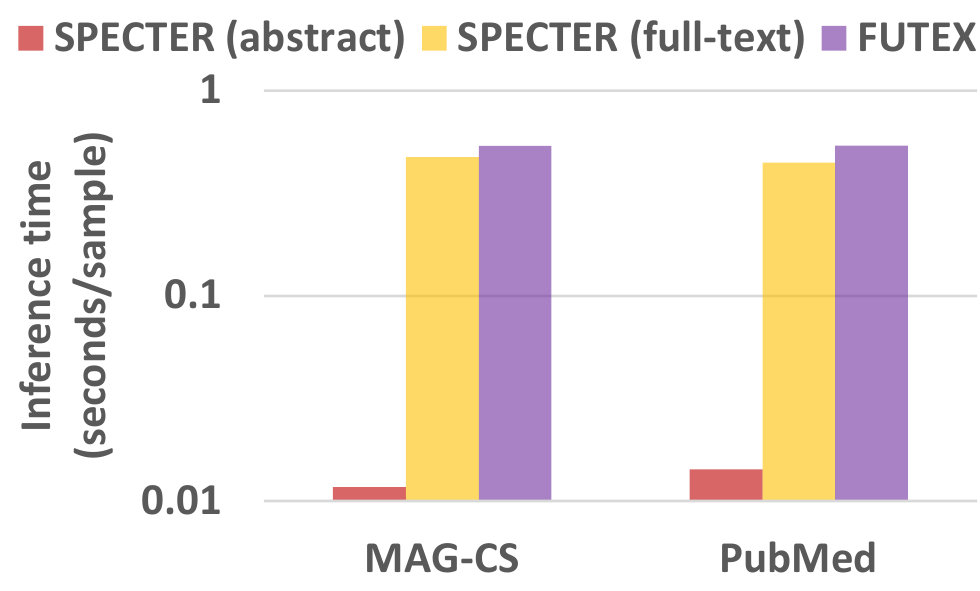}}
\vspace{-1em}
\caption{Training and inference time of \ours and representative baselines on MAG-CS and PubMed.} 
\vspace{-1em}
\label{fig:time}
\end{figure}

We now analyze the training and inference time of \ours on MAG-CS and PubMed. To be specific, we compare the training time of \ours (the network-aware contrastive learning step) with that of PLM+GAT and GraphFormers, which also use network signals for training. For a fair comparison, we run each model on an NVIDIA RTX A6000 GPU, and we train the three models for the same number of epochs on each dataset (i.e., 20 epochs on MAG-CS and 5 epochs on PubMed). As for the inference stage, we notice that the major factor which affects each model’s inference efficiency is whether it is used for abstracts only or full texts. If we fix this factor, different BERT-based baselines will have similar inference efficiency because they have similar model sizes and architectures. Therefore, we choose SPECTER as a representative for comparison and report its inference time (per testing sample) when used for abstracts only and full texts. The results are demonstrated in Figure \ref{fig:time}.

From Figure \ref{fig:time}(a), we observe that \ours has much less training time than PLM+GAT and GraphFormers. From Figure \ref{fig:time}(b), we find that the inference efficiency of \ours is on par with SPECTER (full-text). In comparison with SPECTER (abstract), \ours and SPECTER (full-text) need significantly more time to run because about 30 to 45 times more paragraphs are considered.

\section{Related Work}
\noindent \textbf{Weakly Supervised Text Classification.} Weakly supervised text classification aims to assign relevant label(s) to each document without any human-annotated training samples provided. The common formats of weak supervision include label names \cite{meng2020text}, a small set of category-indicative keywords \cite{mekala2020contextualized}, and label descriptions \cite{zhang2022metadata}. Technically, earlier methods mainly utilize Explicit Semantic Analysis \cite{chang2008importance,song2014dataless}, Latent Dirichlet Allocation \cite{chen2015dataless,li2016effective,li2018dataless}, and context-free word embeddings \cite{meng2018weakly,meng2019weakly}. Inspired by the success of BERT \cite{devlin2019bert} in a wide spectrum of text mining tasks, recent studies start to exploit the power of PLMs in weakly supervised text classification. For example, ConWea \cite{mekala2020contextualized} uses BERT to disambiguate the provided keywords and retrieve more category-indicative words for pseudo training data collection; LOTClass \cite{meng2020text} leverages one BERT encoder to perform masked language modeling for finding more indicative words and another BERT to perform classification; X-Class \cite{wang2021x} uses BERT representations of words to perform category-aware clustering and then aligns documents to categories; LIME \cite{park2022lime} adopts BART-large-MNLI \cite{lewis2020bart} and prompts to predict pseudo labels of each document and then uses BERT for self-training. However, all the aforementioned methods focus on a relatively small label space ($\leq 50$ categories in most cases) and assume each document is relevant to one category only (or a single path from the root to a leaf category in the hierarchical classification setting). In contrast, \ours studies larger and more fine-grained label spaces (e.g., $> 10,000$ categories) where each document is relevant to multiple labels in most cases.

\vspace{1mm}

\noindent \textbf{Zero-Shot Multi-Label Text Classification.} In general, similar to ``weakly supervised'', the term ``zero-shot'' also implies that the classifier does not need any annotated samples. However, in large-scale or extreme multi-label text classification, ``zero-shot'' is interpreted in different ways in the existing literature. According to \cite{yin2019benchmarking}, the \textit{restrictive} zero-shot setting assumes training samples are given for some seen classes and the classifier aims to predict unseen classes \cite{chalkidis2019large,nam2015predicting,nam2016all,rios2018few,gupta2021generalized,xiong2022extreme,zhang2019integrating}, which is different from our ``weakly supervised'' setting; the \textit{wild} zero-shot setting does not assume any seen classes and the classifier needs to make predictions without relying on any annotations \cite{shen2021taxoclass,yin2019benchmarking,zhang2022metadata}. For example, TaxoClass \cite{shen2021taxoclass} uses RoBERT-large-MNLI \cite{liu2019roberta} to convert text classification to an entailment task; MICoL \cite{zhang2022metadata} proposes a metadata-induced contrastive learning method to fine-tune SciBERT \cite{beltagy2019scibert}. However, existing \textit{wild} zero-shot classifiers still view each document as a linear sequence of paragraphs, thus cannot be directly applied to full-text paper classification due to the maximum length limit of PLMs. In comparison, \ours exploits the cross-paper and in-paper structures of scientific literature.

\vspace{1mm}

\noindent \textbf{Scientific Paper Classification.} Classifying scientific papers is a common evaluation task in both text mining (e.g., \cite{beltagy2019scibert,cohan2020specter,liu2022oag,zhang2023pre}) and network mining (e.g., \cite{dong2017metapath2vec,zhang2019heterogeneous,hu2020heterogeneous,jin2023patton}) studies. However, most studies consider coarse-grained paper classification only (e.g., $\leq 50$ categories). To satisfy users' fine-grained interests, Zhang et al. \cite{zhang2021match,zhang2022metadata,zhang2023effect} and Ye et al. \cite{ye2021beyond} propose to use paper metadata to perform large-scale multi-label paper classification. In the biomedical domain, MeSH indexing \cite{liu2015meshlabeler,peng2016deepmesh,xun2019meshprobenet} can also be cast as a multi-label text classification task to tag PubMed papers with fine-grained medical subject headings. Nevertheless, these studies focus on using the paper title and abstract only. To the best of our knowledge, FullMeSH \cite{dai2020fullmesh} and BERTMeSH \cite{you2021bertmesh} are two representative studies making use of paper full texts. However, they adopt a fully supervised setting and are not directly applicable to our task.

\section{Conclusions and Future Work}
We present \ours, a multi-label scientific paper classifier that relies on label names and descriptions as the only supervision and does not require any human-annotated training data. \ours exploits paper full texts and consists of three modules: the contrastive learning module leverages the cross-paper citation network structure; the semantic aggregation module uses the in-paper hierarchy structure of sections, subsections, and paragraphs; the self-training module trains a full-text classifier using pseudo labels to complement the initial predictions. Experiments on two datasets demonstrate the superiority of \ours over competitive weakly supervised baselines and show that \ours is on par with fully supervised classifiers with thousands of ground-truth training samples. An ablation study validates the usefulness of all three proposed modules. A case study shows that \ours can effectively extract signals from full text to predict labels not indicated by the paper title or abstract.

Interesting future directions include (1) how to leverage other in-paper structural signals, such as the relationship between paragraphs and figures/tables, to further improve the classification performance and (2) how to leverage large language models (e.g., GPT-4 \cite{openai2023gpt4}) for weakly supervised fine-grained paper classification, where one needs to tackle the maximum input length limit given the paper full text and tens of thousands of label names.

\section*{Acknowledgments}
We thank anonymous reviewers for their valuable and insightful feedback.
Research was supported in part by the IBM-Illinois Discovery Accelerator Institute, US DARPA KAIROS Program No. FA8750-19-2-1004 and INCAS Program No. HR001121C0165, National Science Foundation IIS-19-56151, IIS-17-41317, and IIS 17-04532, and the Molecule Maker Lab Institute: An AI Research Institutes program supported by NSF under Award No. 2019897, and the Institute for Geospatial Understanding through an Integrative Discovery Environment (I-GUIDE) by NSF under Award No. 2118329. Any opinions, findings, and conclusions or recommendations expressed herein are those of the authors and do not necessarily represent the views, either expressed or implied, of DARPA or the U.S. Government.

\newpage
\bibliographystyle{ACM-Reference-Format}
\bibliography{kdd}

\appendix
\section{Appendix}
\subsection{The Entire Procedure of \ours}
\label{sec:pseudocode}
We summarize the entire procedure of \ours in Algorithm \ref{alg:fute}.

\begin{algorithm}[h]
\caption{\ours}
\label{alg:fute}
\small
\KwIn{
A collection of unlabeled scientific papers $\mathcal{D}$; the cross-paper network structure $\mathcal{G}$; each paper $d$'s full text and its in-paper hierarchy structure $\mathcal{T}_d$; a label space $\mathcal{L}$; each label $l$'s name and description.
}
\KwOut{The relevant labels $\mathcal{L}_d \subseteq \mathcal{L}$ of each paper $d\in \mathcal{D}$.}
Retrieve a small set of candidate labels $\mathcal{C}(d) \subseteq \mathcal{L}$ for each paper using lexical matching\;
\textcolor{myblue}{// Network-Aware Contrastive Fine-Tuning\;}
Fine-tune a PLM using the contrastive loss in Eq. (\ref{eqn:contrastive})\;
\For{$d \in \mathcal{D}$}
{
    \For{$l \in \mathcal{C}(d)$}
    {
        $\score_X(d, l) \gets$ Eq. (\ref{eqn:scorex})\;
    }
}
\textcolor{myblue}{// Hierarchy-Aware Aggregation\;}
\For{$d \in \mathcal{D}$}
{
    \For{$p \in d$}
    {
        $\bmh_p \gets$ Eq. (\ref{eqn:bi})\;
    }
    $\bmh_d \gets$ Eq. (\ref{eqn:aggr}) according to the hierarchy $\mathcal{T}_d$\;
}
\For{$l \in \mathcal{C}(d)$}
{
    $\bmh_l \gets$ Eq. (\ref{eqn:bi})\;
}
\For{$d \in \mathcal{D}$}
{
    \For{$l \in \mathcal{C}(d)$}
    {
        $\score_B(d, l) \gets$ Eq. (\ref{eqn:scoreb})\;
    }
}
${\rm MRR}(l|d) \gets$ Eq. (\ref{eqn:mrr})\;
\textcolor{myblue}{// Self-Training\;}
\For{$d \in \mathcal{D}$}
{
    $\bmx_d \gets$ Eq. (\ref{eqn:selfx})\;
    $\bmy_d \gets$ Eq. (\ref{eqn:selfy})\;
}
Train a Parabel classifier $\Pr(\hat{\bmy}_d|\bmx_d)$ using $\bmx_d$ and $\bmy_d$\;
Get the final ranking list following Example \ref{ex:final}\;
Return $\mathcal{L}_d =\{$top-$k$ ranked labels of $d\}$\;
\end{algorithm}

\subsection{Performance on Infrequent Labels}
\begin{table*}[!t]
\centering
\caption{PSP@$k$ and PSN@$k$ scores of compared methods on MAG-CS and PubMed. Bold, *, and **: the same meaning as in Table \ref{tab:performance}.}
\scalebox{0.9}{
\begin{tabular}{c|ccccc|ccccc}
\hline
\multirow{2}{*}{\textbf{Method}} & \multicolumn{5}{c|}{\textbf{MAG-CS \cite{wang2020microsoft}}}  & \multicolumn{5}{c}{\textbf{PubMed \cite{lu2011pubmed}}}  \\ \cline{2-11}
& \textbf{PSP@1} & \textbf{PSP@3} & \textbf{PSP@5} & \textbf{PSN@3} & \textbf{PSN@5} & \textbf{PSP@1} & \textbf{PSP@3} & \textbf{PSP@5} & \textbf{PSN@3} & \textbf{PSN@5} \\ \hline
SPECTER \cite{cohan2020specter} (abstract)        & 0.5222**         & 0.5529**         & 0.5511**         & 0.5448**         & 0.5450**         & 0.3712**         & 0.3577**         & 0.3427**         & 0.3617**         & 0.3519**         \\
0SHOT-TC \cite{yin2019benchmarking,liu2019roberta}        & 0.4417**         & 0.4648**         & 0.4824**         & 0.4584**         & 0.4696**         & 0.3410**         & 0.3395**         & 0.3324**         & 0.3401**         & 0.3357**         \\
MICoL \cite{zhang2022metadata}           & \textbf{0.5419}         & 0.5675         & 0.5604**         & \textbf{0.5610}         & 0.5579         & \textbf{0.4196}         & 0.3872         & 0.3625**         & \textbf{0.3964}         & 0.3798         \\ \hline
SPECTER \cite{cohan2020specter} (full text)        & 0.5180**         & 0.5521**         & 0.5510**         & 0.5431**         & 0.5438**         & 0.3622**         & 0.3657**         & 0.3530**         & 0.3654**         & 0.3577**         \\
PLM+GAT \cite{yang2021graphformers,velickovic2018graph}     & 0.4920** & 0.5335** & 0.5398** & 0.5225** & 0.5278** & 0.3337** & 0.3364** & 0.3324** & 0.3359** & 0.3335**         \\
GraphFormers \cite{yang2021graphformers}    & 0.4937** & 0.5324** & 0.5380** & 0.5222** & 0.5270** & 0.3460** & 0.3454** & 0.3374** & 0.3459** & 0.3409**         \\ \hline
\ours            & 0.5320         & \textbf{0.5682}         & \textbf{0.5678}         & 0.5588         & \textbf{0.5600}         & 0.4071         & \textbf{0.3882}         & \textbf{0.3764}         & 0.3933         & \textbf{0.3853}   \\ \hline     
\end{tabular}
}
\label{tab:propensity}
\vspace{0.5em}
\end{table*}

A large and fine-grained label space typically implies a long-tailed label distribution, where most categories are associated with only a few documents. In many real applications, it is desirable to predict more tail labels. For example, in scientific paper classification, predicting a paper is relevant to ``\textsf{Lagrangian Support Vector Machine}'' is more informative than saying the paper the relevant to ``\textsf{Machine Learning}''. To promote prediction of tail labels, recent studies \cite{jain2016extreme,wei2021towards,gupta2021generalized,you2019attentionxml,zhang2022metadata} propose to use propensity-based P$@k$ (i.e., PSP$@k$) and propensity-based NDCG$@k$ (i.e., PSN$@k$) as evaluation metrics. PSP$@k$ and PSN$@k$ are formally defined as follows.
\begin{equation}
\small
\begin{split}
    \frac{1}{p_l} = 1+C(N_l+B)^{-A}, \ \ \ {\rm PSP}&{@}k = \frac{1}{k} \sum_{i = 1}^k \frac{z_{d, {\rm rank}(i)}}{p_{{\rm rank}(i)}}. \\
    {\rm PSDCG@}k = \sum_{i=1}^k \frac{z_{d, {\rm rank}(i)}}{p_{{\rm rank}(i)}\log(i+1)}, \ \ \ &{\rm PSN@}k = \frac{{\rm PSDCG@}k}{\sum_{i=1}^{\min(k, ||\bmz_d||_0)}\frac{1}{\log(i+1)}}.
\end{split}
\label{eqn:pspatk}
\end{equation}
The intuition behind these two metrics is to give a higher reward to a model if it predicts an infrequent label correctly. In Eq. (\ref{eqn:pspatk}), $\frac{1}{p_l}$ is such a reward; $N_l$ is number of papers relevant to $l$ in the whole dataset $\mathcal{D}$; $A, B, C > 0$ are constants. In this way, the less frequent a label is, the higher reward a model can get when predicting it correctly. PSP$@k$ and PSN$@k$ scores can be viewed as a reward-weighted version of P$@k$ and NDCG$@k$, respectively. Following previous studies \cite{jain2016extreme,wei2021towards,you2019attentionxml,zhang2022metadata}, we set $A=0.55$, $B=1.5$, and $C=(\log|\mathcal{D}|-1)(B+1)^A$. By definition, we have PSP@1 $\equiv$ PSN@1 if each paper has at least one ground-truth label.

Table \ref{tab:propensity} shows PSP$@k$ and PSN$@k$ scores of \ours and competitive baselines on MAG-CS and PubMed. From Table \ref{tab:propensity}, we can observe that: (1) \ours consistently and significantly outperforms all baselines except MICoL. (2) When comparing with MICoL, \ours has lower PSP@1 and PSN@3 but higher PSP@3, PSP@5, and PSN@5. The only statistically significant gap between \ours and MICoL is the gap of PSP@5. This echos our finding from Table \ref{tab:performance} that considering paper full texts is more beneficial to lower-ranked predictions. One possible reason why \ours underperforms MICoL in terms of PSP@1 and PSN@3 is that \ours ensembles the predictions of a Cross-Encoder (Eq. (\ref{eqn:scorex})) and a Bi-Encoder (Eq. (\ref{eqn:scoreb})) while MICoL is solely based on a Cross-Encoder according to our usage. In fact, as shown in \cite{zhang2022metadata}, labels predicted by the Cross-Encoder architecture are more infrequent than those by the Bi-Encoder.

\subsection{Performance on a Small Dataset}
\begin{table*}[!t]
\caption{P@$k$ and NDCG@$k$ scores of compared methods on the Art dataset. Bold, *, and **: the same meaning as in Table \ref{tab:performance}.}
\scalebox{0.9}{
\begin{tabular}{c|ccccc}
\hline
\multirow{2}{*}{\textbf{Method}} & \multicolumn{5}{c}{\textbf{Art \cite{zhang2023effect}}} \\
\cline{2-6}
& \textbf{P@1}    & \textbf{P@3}    & \textbf{P@5}    & \textbf{NDCG@3} & \textbf{NDCG@5} \\
\hline
SPECTER \cite{cohan2020specter} (abstract)                         & 0.4146**          & 0.2693**          & 0.1890**          & 0.3452**          & 0.3349**          \\
0SHOT-TC \cite{yin2019benchmarking,liu2019roberta}                        & \textbf{0.4573} & 0.2724**          & 0.1927**          & 0.3559**          & 0.3479**          \\
MICoL \cite{zhang2022metadata}                           & 0.4305          & 0.2699**          & 0.1933**          & 0.3510**          & 0.3447**          \\
\hline
SPECTER \cite{cohan2020specter} (full text)                         & 0.4238**          & 0.2713**          & 0.1921**          & 0.3510**          & 0.3432**          \\
\hline
\ours                             & 0.4347          & \textbf{0.3049} & \textbf{0.2211} & \textbf{0.3852} & \textbf{0.3820} \\
\hline
\end{tabular}
}
\label{tab:art}
\end{table*}

MAG-CS and PubMed have nearly 97K and 252K papers, respectively, which can provide rich self-supervision during contrastive learning and self-training. We now examine the performance of \ours on a small dataset and check if it can still outperform competitive baselines. To facilitate this, we adopt the Art dataset from the MAPLE benchmark \cite{zhang2023effect}. These Art papers are labeled with 1,990 categories at different granularities (e.g., ``\textsf{classics}'', ``\textsf{popular music}'', and ``\textsf{rhetorical criticism}''), and we manage to find 328 of them from S2ORC \cite{lo2020s2orc} to obtain full texts. The performance of \ours and competitive baselines on these Art papers are demonstrated in Table \ref{tab:art}.
We can observe that \ours performs the best in terms of P@3, P@5, NDCG@3, and NDCG@5. For P@1, \ours is second to 0SHOT-TC. This observation implies that even if the dataset is small (which may limit the power of contrastive learning and self-training), \ours still works effectively.

\subsection{Hyperparameter Study}
\begin{figure}[!t]
\centering
\subfigure[]{
\includegraphics[width=0.23\textwidth]{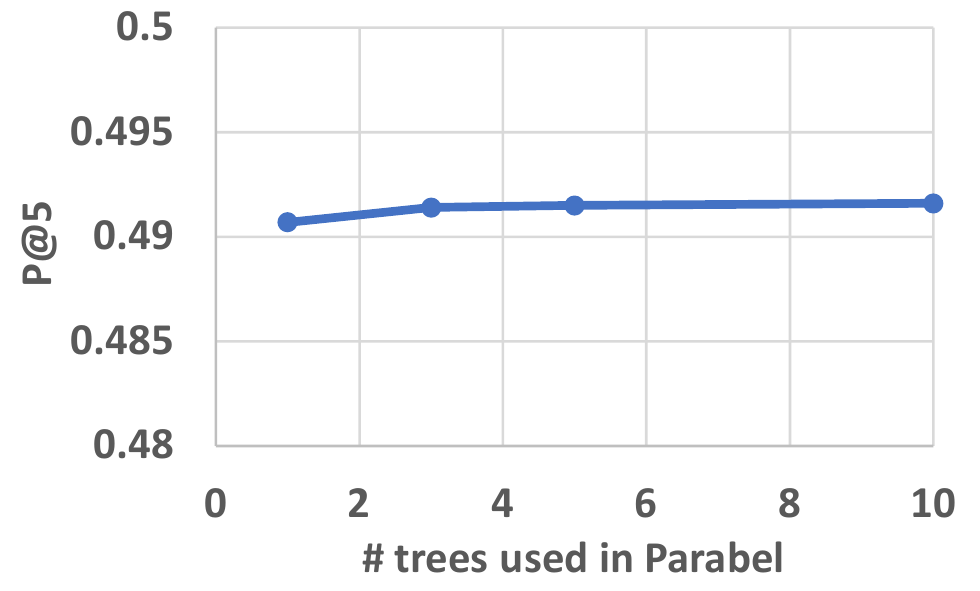}}
\hspace{-1mm}
\subfigure[]{
\includegraphics[width=0.23\textwidth]{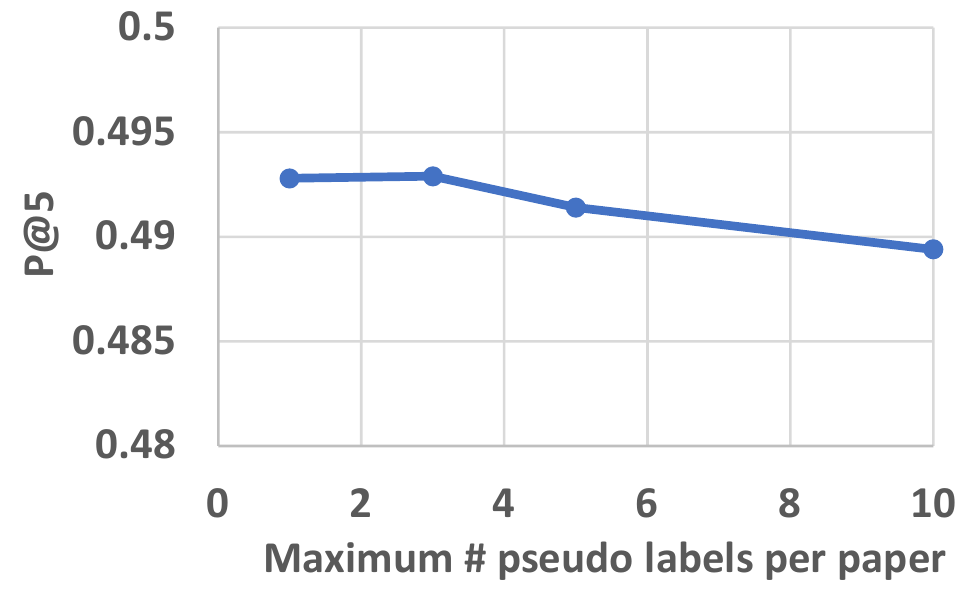}}
\vspace{-1em}
\caption{Parameter sensitivity analysis on MAG-CS. (a) Effect of the number of trees used in the Parabel classifier. (b) Effect of the maximum number of pseudo labels per paper used for self-training.} 
\label{fig:parameter}
\end{figure}

We study the effect of two major hyperparameters in \ours: the number of trees used in the Parabel classifier and the maximum number of pseudo labels per paper used for self-training (i.e., $N$). The P@5 scores of \ours on MAG-CS with different hyperparameter values in $\{1,3,5,10\}$ are plotted in Figure \ref{fig:parameter}.
We can find that the performance of \ours is not quite sensitive to the two hyperparameters. Indeed, all P@5 scores shown in Figure \ref{fig:parameter} outperform those of all baselines in Table \ref{tab:performance}.
\end{spacing}

\end{document}